# Biometric Recognition System (Algorithm)

Rahul Kumar Jaiswal* and Gaurav Saxena**
*The LNM Institute of Information Technology
rahulkumarjaiswal@gmail.com
**Jaypee University of Engineering and Technology
gaurav.saxena@juet.ac.in

**Abstract:** Fingerprints are the most widely deployed form of biometric identification. No two individuals share the same fingerprint because they have unique biometric identifiers. This paper presents an efficient fingerprint verification algorithm which improves matching accuracy. Fingerprint images get degraded and corrupted due to variations in skin and impression conditions. Thus, image enhancement techniques are employed prior to singular point detection and minutiae extraction. Singular point is the point of maximum curvature. It is determined by the normal's of each fingerprint ridge, and then following them inward towards the centre. The local ridge features known as minutiae is extracted using cross-number method to find ridge endings and ridge bifurcations. The proposed algorithm chooses a radius and draws a circle with core point as centre, making fingerprint images rotationally invariant and uniform. The radius can be varied according to the accuracy depending on the particular application. Morphological techniques such as clean, spur and H-break is employed to remove noise, followed by removing spurious minutiae. Templates are created based on feature vector extraction and databases are made for verification and identification for the fingerprint images taken from Fingerprint Verification Competition (FVC2002). Minimum Euclidean distance is calculated between saved template and the test fingerprint image template and compared with the set threshold for matching decision. For the performance evaluation of the proposed algorithm various measures, equal error rate (EER), $D_{min}$ at EER, accuracy and threshold are evaluated and plotted. The measures demonstrate that the proposed algorithm is more effective and robust.

## Introduction

Biometric is the science of automatically and uniquely recognizing individuals based upon one or more instinctive physical or behavioral characteristics [12]. Present day biometric systems are becoming an essential component of effective personal identification. Fingerprints are the most widely adopted tool for personal identification among all biometrics due to easily acceptance, distinctiveness and permanence. Despite the ingenious methods improvised to increase the efficiency of the manual approach to fingerprint indexing and matching, the ever growing demands on fingerprint recognition quickly became overwhelming. Therefore, fingerprint identification is one of the main motives of various research endeavors in the field of Pattern recognition over the last few decades. The principle area of fingerprint identification includes law enforcement, criminal identification, fraud prevention, and computer access etc [15].

The main purpose of biometric recognition is to minimize the degree of forgery of the desired system and to improve one or more perceptual aspects of identification, such as the quality and/or matching accuracy so that the required service will be accessed by the legitimate user only [20]. A fingerprint is a unique pattern of ridges and valleys on the surface of a finger of an individual. A ridge is defined as a single curved segment, and a valley is the region between two adjacent ridges. The classification of fingerprint is based on micro and macro features [13]. Macro features include ridge patterns like loop, whorl and arc, along with core point, delta point and ridge count. Core point is found at the center of finger image and it is the uppermost of the innermost curving ridge whereas delta point is the meeting point of two or three ridges. Micro features include minutiae, it's orientation and position. Minutiae points are the local ridge discontinuities. These are of two types namely; terminations, which is the immediate ending of a ridge and bifurcations, which is the point on the ridge from which two branches derive [1]. A good quality image has around 40 to 100 minutiae [5]. Minutiae points are used for determining uniqueness of a fingerprint. Orientation refers to the direction in which a minutiae feature appears to be moving and position implies the relative location of the minutiae. The micro and macro features are shown in Fig. 1 and Fig. 2.

Fingerprint recognition system is categorized as enrollment, verification, and identification process system [14]. Enrollment includes capturing image, extracting feature, creating templates and making database. Verification authenticates a person's identity by comparing captured biometric traits with previously enrolled biometric reference template, pre-stored in the system. It conducts one-to-one comparison. Identification or Authentication recognizes an individual by searching the entire enrolled template database for a match. It conducts one-to-many comparisons. Authentication can be understood as a combination of four RIGHTS [20].

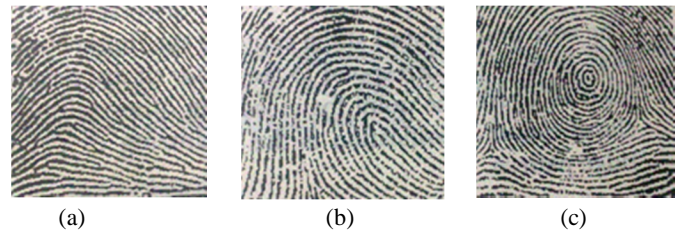

Fig. 1. Macro features: ridge pattern (a) arc (b) loop (c) whorl

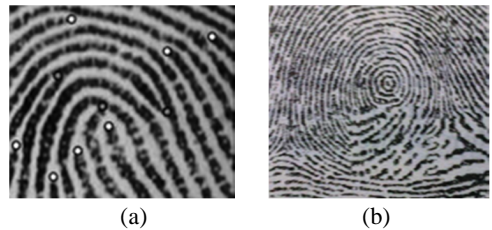

Fig. 2. Micro features: minutiae points (a) terminations (b) bifurcations

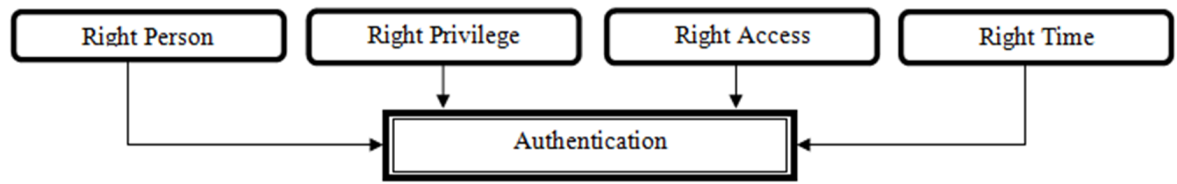

Fig. 3. Components of authentication

In this paper, Fingerprint matching system based on minutiae extraction is proposed to enhance the accuracy of the biometric system degraded by elements of noise and large variability in different impressions of same finger [4]. The minutiae points are determined by cross number [8, 21]. The cross number is 1 for termination points and for bifurcation points, it is 3. Our first part of the problem is to identify fingerprints based on feature extraction. After getting minutiae as feature vector in region of interest (ROI), templates of fingerprint database are created. The minimum Euclidean distance is determined and based on certain threshold, matching will be decided with the database saved templates [7]. The proposed recognition algorithm allows in finding the best tradeoff between the radius of circle, matching accuracy and the space requirements in selection of radius in a perceptive view.

The paper is systematized as follows. In Section 2, the review of Pattern based fingerprint recognition method is described with its connection in fingerprint matching. In Section 3, the Minutiae based finger print recognition approach is described. In Section 4, the simulation results, performance measures and analysis of proposed algorithm is detailed. Finally, paper is concluded in Section 5.

## Pattern based Fingerprint Recognition Method:

Pattern based fingerprint recognition method compares the basic fingerprint patterns (arch, loop, and whorl) between a previously stored template and a test fingerprint. The necessity of this method is that the images should be aligned in the same orientation. To do this, the algorithm detects a central point in the fingerprint image and centers on that point. In this method, the template contains the type, size, and orientation of patterns within the aligned fingerprint image. The test fingerprint image is graphically compared with the previously stored template to determine the degree of match [3]. The major disadvantage of pattern based fingerprint recognition method is that they are sensitive to proper placement of finger and need large storage for templates [7]. Several variations of pattern based fingerprint recognition method have been proposed to overcome this problem. The pattern of fingerprint [16-18] is shown in Fig. 1.

## Minutiae based Fingerprint Recognition Algorithm

The minutiae based fingerprint recognition algorithm is relatively stable, robust to contrast, image resolutions, and global deformation as compared to pattern based fingerprint recognition method [6]. This approach is the backbone of the current available fingerprint recognition system. Fingerprint identification with minutiae extraction is mainly based on the minutiae points i.e. the direction and location of the ridge endings and bifurcations along the ridge path [11]. It reduces the complex fingerprint recognition problem to a point pattern matching problem. The proposed algorithm includes pre-processing of fingerprint image, feature extraction, post-processing and finally matching decision. Essentially, the matching consists of finding the minimum difference of distance ($D_{min}$) between the saved template and the test minutiae sets having maximum number of minutiae pairings. The test fingerprint matches if ($D_{min}$) is lower than the set threshold.

**Step-by-step description of Algorithm:**
Step 1: Input the fingerprint image, $f(x, y)$.
Step 2: Conversion of image $f(x, y)$ into grayscale image, $f_g(x, y)$.
Step 3: Resizing image $f_g(x, y)$ to 400 x 400, new image $f_r(x, y)$.
Step 4: Enhancing image using histogram equalization and wiener filter to improve quality, degraded by noise like smudgy area, break in ridge, wounds and sweat. The histogram of a digital image with gray levels in the range [0, L-1] is a discrete function, defined as [15]

$$h(r_k) = n_k \tag{1}$$

$r_k$ - $k^{th}$ Gray level,
$n_k$ - Number of pixels in the image.

Step 5: Finding the core point of the fingerprint image $f_r(x, y)$.
The image is divided into a non-overlapping blocks of size 'w' 10 x 10. The horizontal gradient '$G_x(x, y)$' and vertical gradient '$G_y(x, y)$' at each pixel $(x, y)$ is computed using Sobel mask of size 3 x 3 and the ridge orientation '$\theta(x, y)$' of each pixel is given by [2],

$$G_{xx} = \sum_{(x,y)\epsilon w} G_x^2(x, y) \tag{2}$$

$$G_{yy} = \sum_{(x,y)\epsilon w} G_y^2(x, y) \tag{3}$$

$$G_{xy} = \sum_{(x,y)\in w} G_x(x, y).G_y(x, y) \tag{4}$$

$$\theta(x, y) = \frac{1}{2}\tan^{-1}\left(\frac{2G_{xy}}{G_{xx}-G_{yy}}\right) \tag{5}$$

Now, ridge orientation is smoothed using Gaussian low pass filter. As singular point has the maximum curvature. so, it is located by measuring strength of the peak. Further, applying thinning followed by Morphological closing and opening to locate singular point in original fingerprint image.

Step 6: Extraction of a circle of radius 'R' with core point as centre of the fingerprint image $f_r(x, y)$ to get new image $f_c(x, y)$ in the region of interest (ROI) because area near singular point contains correct and efficient information about fingerprint.

Step 7: Conversion of image $f_c(x, y)$ into binary image $f_b(x, y)$ by thresholding [10]. Pixel value above the threshold is assigned to 1 and below to 0. Here threshold = 160.

Step 8: Applying thinning operation on the image $f_b(x, y)$ to get thinned image $f_t(x, y)$. Thin operation reduces width of ridges to one pixel wide.

Step 9: Extracting minutiae points (terminations and bifurcations) of $f_t(x, y)$ using Cross-number (CN) concept [8]. It is computationally efficient and inherently simple. The minutiae points are extracted by scanning the local neighborhood of each pixel in the ridge thinned image, using a 3 x 3 window (Fig. 4).

| a) | $P_4$ | $P_3$ | $P_2$ | b) | 1 | **0** | 1 | c) | 1 | **0** | 1 |
|---|---|---|---|---|---|---|---|---|---|---|---|
| | $P_5$ | **(x,y)** | $P_1$ | | 1 | **(x,y)** | 1 | | 1 | **(x,y)** | 0 |
| | $P_6$ | $P_7$ | $P_8$ | | 1 | 1 | 1 | | 1 | **0** | 1 |

Fig 4. a) 3 × 3 window b) Ridge ending and c) bifurcation

The CN value [8] is defined as half the sum of the differences between pairs of adjacent pixels, $P_i$ and $P_{i+1}$ in the eight neighborhood and computed by,

$$CN_{(x,y)} = \frac{1}{2} \sum_{i=1}^{8} |P_i - P_{i+1}|, \quad P_9 = P_1 \qquad (6)$$

The ridge pixel is classified as a ridge ending and bifurcation having cross-number 1 and 3 respectively. Data matrix is generated to get the position, orientation and type of minutiae.

Step 10: Post-processing to remove spurious minutiae, observed due to undesired spikes, breaks, and holes. Morphological operation [9] namely clean, spur and H-break is employed on thinned image $f_t(x,y)$ to get image $f_m(x,y)$ as described,

Clean:
$$\begin{matrix} 0 & 0 & 0 \\ 0 & 1 & 0 \\ 0 & 0 & 0 \end{matrix} \text{ becomes } \begin{matrix} 0 & 0 & 0 \\ 0 & 0 & 0 \\ 0 & 0 & 0 \end{matrix} \text{ after clean operation}$$

Spur:
$$\begin{matrix} 0 & 0 & 0 \\ 0 & 1 & 0 \\ 1 & 0 & 0 \end{matrix} \text{ becomes } \begin{matrix} 0 & 0 & 0 \\ 0 & 0 & 0 \\ 1 & 0 & 0 \end{matrix} \text{ after spur operation}$$

H-Break:
$$\begin{matrix} 1 & 1 & 1 \\ 0 & 1 & 0 \\ 1 & 1 & 1 \end{matrix} \text{ becomes } \begin{matrix} 1 & 1 & 1 \\ 0 & 0 & 0 \\ 1 & 1 & 1 \end{matrix} \text{ after H-Break operation}$$

Step 11: Finding true minutiae points in ROI of $f_m(x,y)$ to get final image $f_{final}(x,y)$ after removing spurious minutiae [14] in the cases, if
i) distance between a termination and a bifurcation is smaller than D
ii) distance between two bifurcations is smaller than D
iii) distance between two terminations is smaller than D
'D' is the average distance between minutiae points. Here D = 6.

Step 12: Representation of linear distance and angle of each minutia in ROI with respect to core point in polar form. The linear distance and angle between core point $(x_1, y_1)$ and minutia $(x_2, y_2)$ is given by,

$$r = D(x,y) = \sqrt{(x_2 - x_1)^2 + (y_2 - y_1)^2} \qquad (7)$$

$$\theta = \tan^{-1}\left(\frac{y_2 - y_1}{x_2 - x_1}\right) \qquad (8)$$

$$z = re^{i\theta} \qquad (9)$$

Step 13: Taking Fourier transform of (9) and saving the Fourier coefficients in '.dat' file.
Step 14: Creation of template of fingerprint database.
Step 15: Calculation of the parameter Euclidean distance ($D_{min}$) between saved template and the test fingerprint template using (7).
Step 16: Minimum ($D_{min}$) is compared with the set threshold to get result whether 'match' or 'not match'.
The block diagram of the minutiae based fingerprint recognition algorithm is shown in Fig. 5.

## Simulation and Results Analysis:

This section presents the simulation results and performance evaluation of the proposed algorithm, minutiae based fingerprint recognition. For simulations, we have employed MATLab software as the simulation platform. For the experimental purpose, the fingerprint samples have been taken from Fingerprint Verification Competition (FVC2002) [22], which is a publicly available fingerprint database and are usually used for benchmark experiments. FVC2002 comprises of four different image sized fingerprint database, collected by four different sensors or technologies. We have employed first database (DB1) due to good quality and image size of fingerprint samples. The resolution of fingerprint image is set to 500 dpi. We have taken gray scale fingerprint image samples for processing.

Each fingerprint images have different and unique characteristics and therefore present a different impact on recognition system. Thus, to achieve a good number of minutiae pairings resulting in maximum accuracy, an appropriate radius of circle is drawn and processed as seen in Table 1. Series of experiments have been performed at different radius value for the test fingerprint samples and the templates are tested with the pre-stored saved template database.

In our experiments, the image size of fingerprint samples is chosen to be 400 x 400. The samples have been contrasted and enhanced with Histogram equalization and Wiener filter. After getting singular point and minutiae extraction, morphological operations have been used to remove false minutiae to get high accuracy.

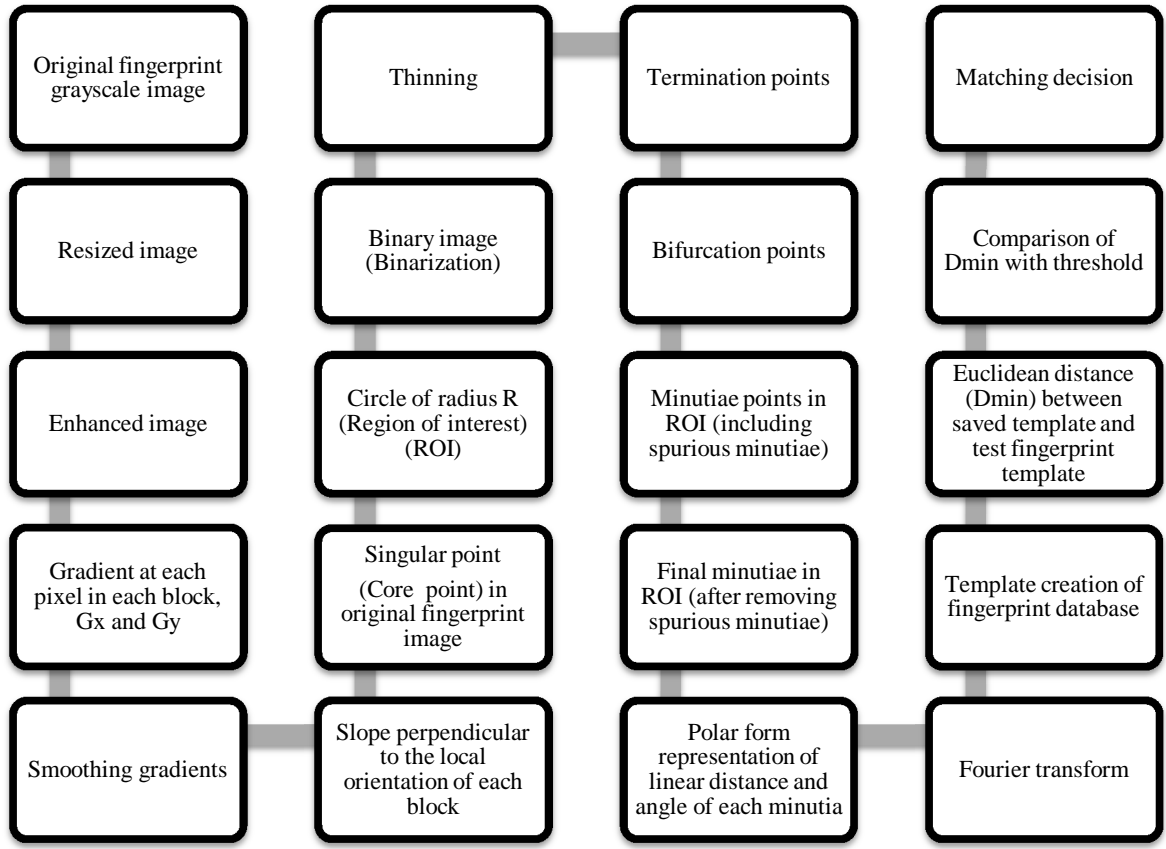

Fig. 5. Block diagram of proposed algorithm

To test the performance of our proposed fingerprint recognition algorithm, the objective measurements like FAR, FRR, EER, $D_{min}$ at EER, and Accuracy have been used. A false acceptance occurs when two images from different fingers are matched, and a false rejection occurs when two images from the same finger are not matched [19]. At equal error rate, both acceptance and rejection errors are equal and it is used to compare the accuracy of biometric system. The EER is calculated using the graph between FAR and FRR versus threshold, which lies in the expected range. The calculated values of these measuring parameters have been tabulated in Table 2.

Further, the step-by-step simulation results of the minutiae based fingerprint recognition algorithm is shown in fig. 6. The FAR, FRR, EER and accuracy with probability and threshold at different radius and same Fourier descriptors are shown in fig. 7.

Table 1. Radius of circle 'R' vs. minutiae points

| S. No. | Radius of circle 'R' | Number of minutiae points in circle of radius 'R' | |
|---|---|---|---|
| | | Fingerprint image sample 1 | Fingerprint image sample 2 |
| 1. | 90 | 75 | 95 |
| 2. | 100 | 84 | 120 |
| 3. | 150 | 186 | 256 |

From the extensive study, it is observed that for every increase in radius of circle (discussed in this paper), the value of input feature vectors or minutiae points of fingerprint image increases and the number of minutiae points is better for fingerprint image samples to identify legitimate user.

The comparative results Equal error rate (EER), $D_{min}$ at EER, Accuracy and Threshold, shown in Table 2, is for different values of radius of circle 'R' and different values of Fourier descriptors. The accuracy gradually increases as the radius of circle 'R' increases, at the same Fourier descriptors level. This means, we can vary radius 'R' according to the accuracy as required in particular application. The EER is minimum and has slight variation for the increasing value of radius of circle. The distance, $D_{min}$ at EER has different values for different values of radius. There is no clear relation between the radius of circle and value of $D_{min}$ due to non coherence but the minimum possible distance is considered for comparison. The overall accuracy is better for our proposed algorithm.

Table 2. Variations of various deciding parameters

| S. No. | Radius of circle 'R' | Fourier descriptors (Real) | Equal error rate (EER) | $D_{min}$ at EER | Accuracy |
|---|---|---|---|---|---|
| 1. | 90 | 80 | 3.00 % | 420 | 94.00 % |
| 2. | 90 | 120 | 3.19 % | 1290 | 93.61 % |
| 3. | 100 | 80 | 2.90 % | 350 | 94.29 % |
| 4. | 100 | 120 | 3.19 % | 1570 | 93.86 % |
| 5. | 150 | 80 | 3.09 % | 290 | 94.81 % |
| 6. | 150 | 120 | 3.18 % | 450 | 93.97% |

The simulation results of the minutiae based fingerprint recognition algorithm is shown in Fig. 6.

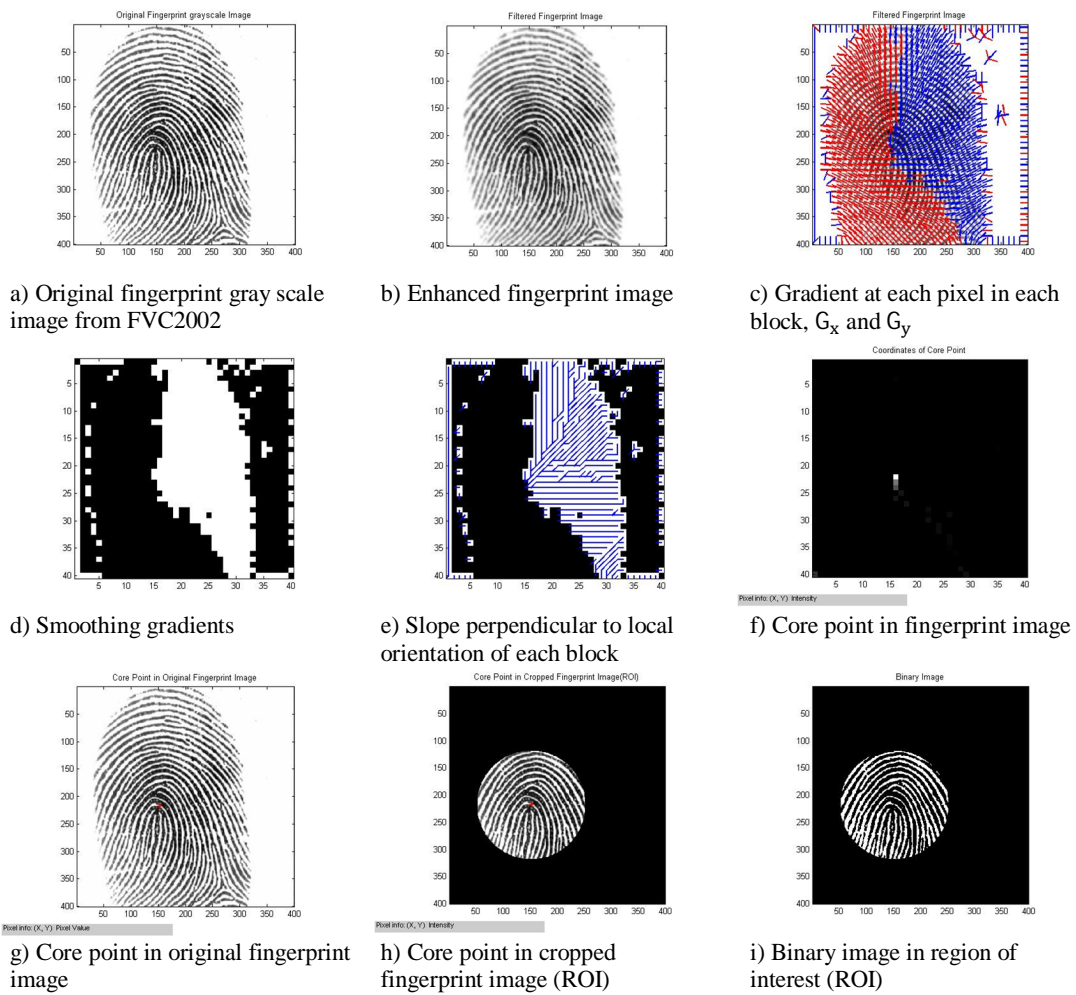

a) Original fingerprint gray scale image from FVC2002

b) Enhanced fingerprint image

c) Gradient at each pixel in each block, $G_x$ and $G_y$

d) Smoothing gradients

e) Slope perpendicular to local orientation of each block

f) Core point in fingerprint image

g) Core point in original fingerprint image

h) Core point in cropped fingerprint image (ROI)

i) Binary image in region of interest (ROI)

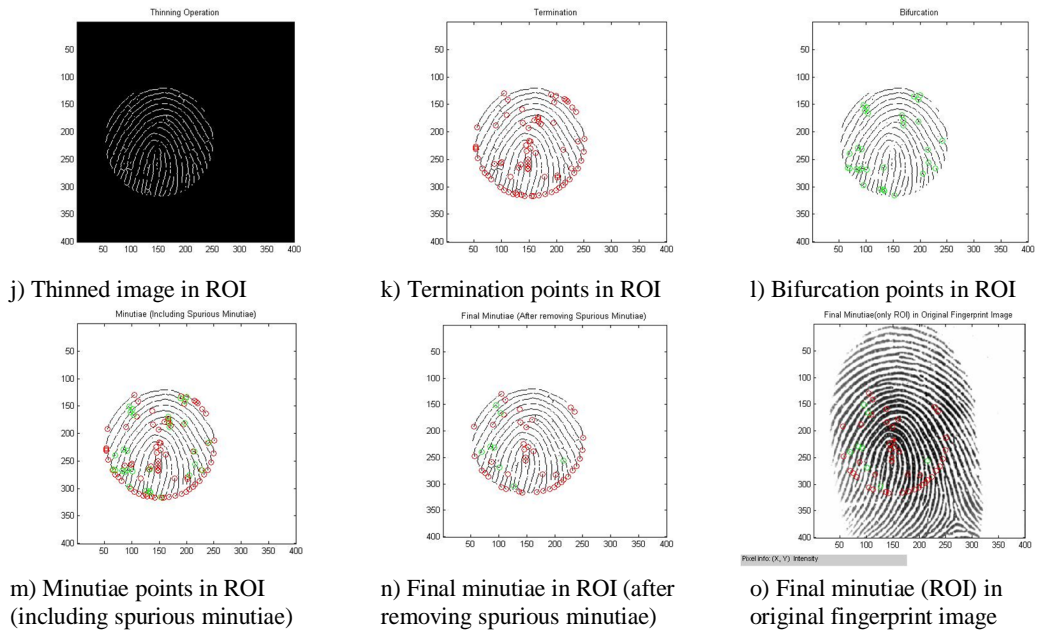

j) Thinned image in ROI
k) Termination points in ROI
l) Bifurcation points in ROI

m) Minutiae points in ROI (including spurious minutiae)
n) Final minutiae in ROI (after removing spurious minutiae)
o) Final minutiae (ROI) in original fingerprint image

Fig 6. Simulation results of the proposed minutiae based fingerprint recognition algorithm

The graph between Probability and Threshold showing FAR, FRR, index of EER and Accuracy at same Fourier descriptors but different Radius is depicted in Fig. 7.

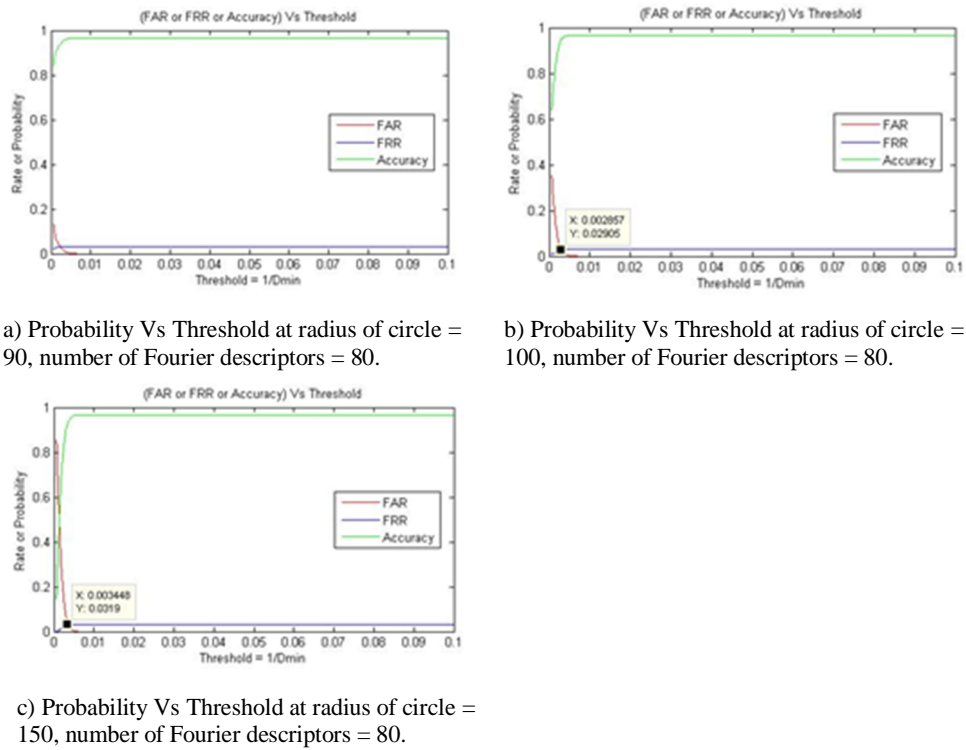

a) Probability Vs Threshold at radius of circle = 90, number of Fourier descriptors = 80.
b) Probability Vs Threshold at radius of circle = 100, number of Fourier descriptors = 80.

c) Probability Vs Threshold at radius of circle = 150, number of Fourier descriptors = 80.

Fig 7. Graph between probability and threshold showing FAR, FRR, index of EER and Accuracy

## Conclusion

In this paper, the Minutiae based Fingerprint Recognition algorithm was implemented to deal the accuracy of fingerprint matching. For the performance evaluation of the algorithm, different thresholds are set to obtain different level of accuracy.

The objective measures were conducted at different values of radius for the fingerprint images. Results show that the quality and accuracy of matched fingerprint enhanced by our proposed algorithm is good while the EER of fingerprint image remains acceptable. Also, the experimental results reveal the effectiveness and robustness of our algorithm but still very few spurious minutiae are left uneliminated. So, there is a scope of developing new techniques which could remove them completely.